# An On-chip Trainable and Clock-less Spiking Neural Network with 1R Memristive Synapses

Aditya Shukla and Udayan Ganguly, *Member, IEEE*

*Abstract*— Spiking neural networks (SNNs) are being explored in an attempt to mimic brain's capability to learn and recognize at low power. Crossbar architecture with highly scalable Resistive RAM or RRAM array serving as synaptic weights and neuronal drivers in the periphery is an attractive option for SNN. Recognition (akin to "reading" the synaptic weight) requires *small* amplitude bias applied across the RRAM to minimize conductance change. Learning (akin to "writing" or updating the synaptic weight) requires *large* amplitude bias pulses to produce a conductance change. The contradictory bias amplitude requirement to perform reading and writing *simultaneously* and *asynchronously*, akin to biology, is a major challenge. Solutions suggested in the literature rely on time-division-multiplexing of read and write operations based on clocks, or approximations ignoring the reading when coincidental with writing. In this work, we overcome this challenge and present a clock-less approach wherein reading and writing are performed in different frequency domains. This enables learning and recognition simultaneously on an SNN. We validate our scheme in SPICE circuit simulator by translating a two-layered feed-forward Iris classifying SNN to demonstrate software-equivalent performance. The system performance is not adversely affected by a voltage dependence of conductance in realistic RRAMs, despite departing from linearity. Overall, our approach enables direct implementation of biological SNN algorithms in hardware.

## I. INTRODUCTION

Since the last century, intense effort has been put into building brain-mimicking electronic systems that can perform intelligent tasks. These tasks include but are not restricted to perception, decision-making, real-time learning and recognition. Deeply inspired from brain, a fairly recent approach is to use spiking neural networks (SNNs) to compute using spatio-temporal spiking events [1]–[3]. Powerful von-Neumann computers [4], [5] suffer from the classical "inter-connect bottleneck" between spatially separated memory and logic units [6]. This inter-connect bottleneck is resolved by a crossbar architecture of memory array addressed by peripheral circuits [7]. Such architectures have been implemented with digital silicon neurons in the periphery with an array of embedded SRAM acting as synapse [8]–[11]. While SRAMs use large (~$800F^2$) cell-size, a typical memristors (or RRAMs) with $4F^2$ cell size leads to significantly denser crossbar arrays [7] (Figure 1a). In the neural network, two operations occur – (a) recognition and (b) learning. For recognition, the conductance of the RRAM simply acts as synaptic weights to convert pre-synaptic neuron spikes into current input for post-synaptic neurons to produce spikes.

For learning, synaptic time dependent plasticity (STDP) based learning rule is used, where the change in weight ($\Delta G$) depends on the difference of spiking-times ($\Delta t$) of a pre- and post-synaptic neuron from the recognition process as shown in Fig. 1b [12]. STDP rule is implemented across a synapse via action potentials from pre- and post-synaptic neuron (Fig. 1c). When a voltage spike is applied on the RRAM, the conductance change ($\Delta G$) of an RRAM depends on peak voltage ($V_{peak}$) i.e. $\Delta G(V_{peak})$ (Fig. 1d). Conductance remains unchanged ($\Delta G = 0$) at low bias. Conductance decreases ($\Delta G < 0$) at a bias exceeding positive threshold ($V > V_{thp}$) and increases ($\Delta G > 0$) as bias decreases below a negative threshold ($V > V_{thp}$). The pre- vs post-neuron spike times (Fig. 1e) are converted to pre- and post-synaptic waveforms (Fig. 1f), which superpose on the RRAM (Fig. 1g) such that the peak voltage depends upon the spike time difference ($\Delta t$) i.e. $V_{peak}(\Delta t)$. As the peak voltage exceeds the threshold of resistive switching in the RRAM, we observe a synaptic conductance change (Fig. 1h). The conductance change is commensurate to peak bias that depends upon $\Delta t$ i.e. $\Delta G(V_{peak}(\Delta t)) = \Delta G(\Delta t)$. Such STDP has been demonstrated in various RRAM devices [13]–[15].

The recognition process is akin to reading, where the synapse conductance is "read" with a *small* bias without altering it, while the learning process is akin to "write", where the synaptic conductivity is modified by a *large* bias. As the post-synaptic neural spikes created during the recognition process are also needed for learning, learning requires recognition to occur simultaneously (Fig. 1i). Here, a "read-write" dilemma occurs due to the contradictory bias requirement of both processes occurring simultaneously, where each process disturbs the other.

In biology, recognition occurs as pre-neuron spikes are converted to post-synaptic current (electrical signal) with a typical timescale of 100ms [16]. For learning, long-term potentiation (LTP) and long-term depression (LTD) are based on pre-/post-neuron spike-time's correlations, also around 100ms [16]. Complex bio-chemical processes enable

This paragraph of the first footnote will contain the date on which you submitted your paper for review. It will also contain support information, including sponsor and financial support acknowledgment..

The authors are with the Department of Electrical Engineering, Indian Institute of Technology Bombay, Mumbai, India – 400076 (e-mail: adityashukla@ee.iitb.ac.in, udayan@ee.iitb.ac.in).

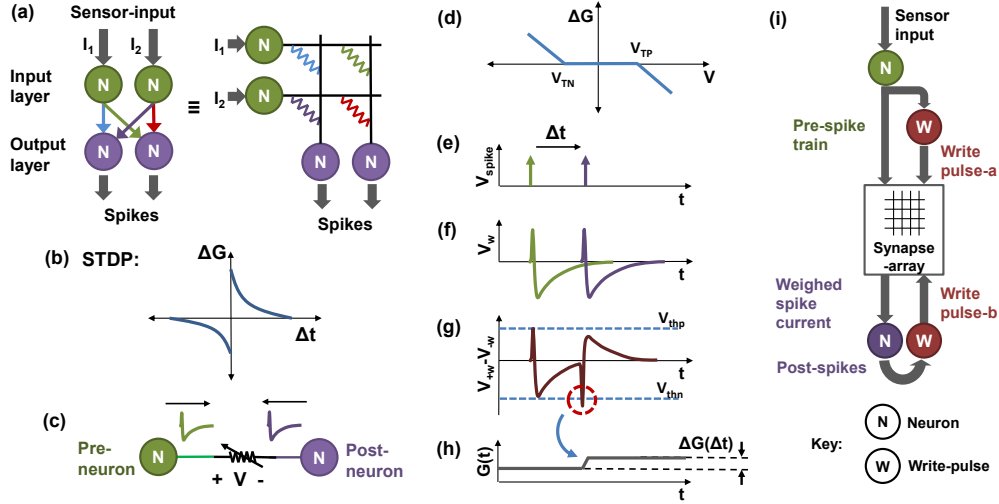

Figure 1: (a) A 2-layer (2×2) SNN compared to an equivalent crossbar implementation; (b) Synaptic-time dependent plasticity (STDP) learning rule (c) Write-pulse are applied from two sides of the RRAM; (d) For an ideal RRAM, the conductance change is proportional to the extent by which pulse's height exceeds the threshold; below the threshold, no change occurs. (e) Pre- and post-synaptic neuron's spikes (green and purple respectively), (f) corresponding waveforms, (g) their superposition, and (h) corresponding change in weight of synapse. (i) High level schematic of a RRAM crossbar based electronic SNN requires simultaneous read and write operations leading to read-write dilemma.

simultaneous and asynchronous process of recognition and learning with similar timescales in biology. On the other hand, contradictory voltage amplitudes for simultaneous recognition and learning operation produces a read-write dilemma in RRAM crossbar array based SNNs. Three approaches have been reported to mitigate this dilemma. First, reading and writing were time-multiplexed using different phases of a clock [11], [17], [18]. Second, an asynchronous approach necessitated switching off the reading process when writing occurs [14], [19]–[21], which significantly affects the general equivalence of hardware to software model during hardware translation. Third, we have suggested a spatial separation of read and write process into two arrays, one for reading and one for writing, to enable simultaneous read and write [22]. This requires that the time-evolving write array's weights be periodically updated into the read array, which is at least twice area inefficient.

In this paper, we propose a scheme inspired from frequency division multiplexing (FDM) to overcome the read-write dilemma. FDM is commonly used for parallel communication of information from multiple independent sources, each using a distinct frequency band.

The paper is organized as follows. In Section II, we set up a general software framework for a basic SNN. In Section III, we present our proposal to solve the read-write dilemma using FDM. In Section IV, we report the SPICE implementation of the SNN for an Iris classifier network in a circuit simulator (SPICE) to demonstrate the algorithmic-equivalence of FDM based hardware solution for "ideal" synapses with linear I-V characteristics. Then, we also present the effects of two non-idealities in realistic synaptic RRAMs – (1) non-ideal resistive switching i.e. dependence of write-voltage thresholds on the conductance-state and (2) non-ideal conductance, i.e. non-linearity in I-V characteristics at read voltages that precedes resistive switching at high voltages. We show that the circuit and software response to the non-ideality is identical. Further, non-ideal conductance based synapses do not degrade the performance of the network. This demonstrates the applicability of FDM to various types of non-linear I-V characteristics of realistic RRAM based synapses.

II. SPIKING NEURAL NETWORK – SOFTWARE FRAMEWORK

We now briefly describe the two essential components of an SNN, taking a two-layered feed-forward network with $m$ input neurons connected to $n$ output neurons through on $m \times n$ synaptic array (Figure 1a). This description is based on an SNN described in detail in [23]. This network shows excellent performance for 4 different classification problems – Fisher Iris dataset, Wisconsin's Breast-Cancer dataset, Wine dataset and Stat-log. All parameters in the following equations used are given in Table 1.

*1) Neuron*

We implement the leaky-integrate and fire (LIF) neuron [3]. The dynamics of the LIF neuron is described by Eq. 1, where, $V(t)$ is the membrane potential, conductance ($G$) and capacitance ($C$) are neuron-specific parameters and $I_i(t)$, $i = 1,2..n$ are the $i^{th}$ pre-synaptic currents. Whenever $V(t)$ reaches a constant threshold ($V_t$), the neuron issues a spike, followed by reset of $V(t)$ to a reset potential $V_R$. The LIF remains at reset potential for a *refractory period* ($\tau_{ref}$) ignoring any input during this time.

$$C \frac{dV_j}{dt} = -GV_j(t) + \sum_i I_{\alpha, i \to j}(t) \quad (1)$$

If $V_j(t') \geq V_T$ then $V(t' \leq t \leq t + \tau_{ref}) = V_R$

*2) Synapse and synaptic time-dependent plasticity*

Synapse generates a time varying current in response to its pre-synaptic neuron's spike, which may excite/inhibit spiking of the post-synaptic neuron (Eq. 2). This current is proportional to the weight ($G_{ij}$) of a synapse between i[th] pre-neuron and j[th] post neuron.

TABLE I
PARAMETERS USED IN SOFTWARE-FRAMEWORK

| Component/Signal | Parameter | Value |
|---|---|---|
| LIF neuron | $C$ | 300 pF |
| | $G$ | 30 nS |
| | $\tau_R$ | 5ms, 0ms |
| | $V_{th}, V_R$ | 90mV, 0 |
| Synapse | $A_+, A_-$ | 9, -15 |
| | $w_{max}, w_{min}$ | 700 S, 0 |
| | $\tau_+, \tau_-$ | 10ms, 20ms |
| | $p$ | 1.7 |
| $\alpha$-function | $\tau_1, \tau_2$ | 2ms, 10ms |
| | $V_0$ | 10pV |

$$I_{\alpha,i \to j}(t) = G_{ij}\alpha_{\tau_1,\tau_2}(t) \quad (2)$$

where,

$$\alpha_{\tau_1,\tau_2}(t) = V_0\left(\exp\left(-\frac{t}{\tau_1}\right) - \exp\left(-\frac{t}{\tau_2}\right)\right) \quad (3)$$

Here, $V_0, \tau_1$ and $\tau_2$ are network design parameters. For learning, the synaptic weight changes according to STDP where weight/conductance change ($\Delta G$) Depending on the time-difference ($\Delta t = t_{post} - t_{pre}$) of the spikes of pre-synaptic ($t_{pre}$) and post-synaptic ($t_{post}$) neurons. Here, STDP rule given in Eq. 3 has been used.

$$\Delta G = \begin{cases} |A_+|\exp\left(-\frac{\Delta t}{\tau_+}\right)\left(1 - \frac{G}{G_{max}}\right)^p, & \Delta t \geq 0 \\ -|A_-|\exp\left(-\frac{\Delta t}{\tau_-}\right)\left(\frac{G}{G_{max}}\right)^p, & \Delta t < 0 \end{cases} \quad (4)$$

Here, $A_+$, $A_-$, $\tau_+$ and $\tau_-$ are network design parameters.

### III. PROPOSAL FOR SOLVING SIMULTANEOUS READ-WRITE DILEMMA USING FDM

To perform read and write concurrently on a RRAM array, we propose a technique inspired from frequency division multiplexing (FDM) used frequently in communication systems. We describe the implementation in three steps: *(A)* reduction of read-pulse width, *(B)* Use of high-frequency sinusoidal spikes for reading and *(C)* Selective read-current extraction.

#### A. Reduction of read-pulse's width

Assuming that the weights evolve slowly (e.g. less than 1% change within a read-pulse [23]), we may treat the weighing process to be linear and time-invariant (LTI; more accurately, *quasi*-LTI). The conversion of pre-synaptic spikes into post-synaptic current (given by Eq. 2) is implemented in circuits using an LTI process consisting of two steps. First, the spike generates a $\alpha$-function voltage. Second, the $\alpha$-function voltage produces a current response proportional to synaptic weight $G_{ij}$ (i.e. its conductance; Figure 2a). Two LTI systems connected in series can be inter-changed without affecting the input-output behavior of the system [24]. Thus, we place the read-pulse generator after (instead of before) the RRAM array (Figure 2b). This enables short duration pre-synaptic spikes in crossbar array instead of extended $\alpha$-functions to reduce the possibility of disturbance to writing process and vice versa, by time overlap of read and write-pulses.

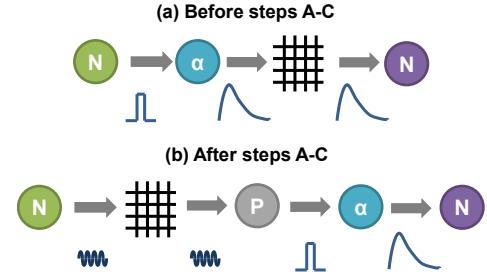

Figure 2: Components' (involved in reading) order before and after applying modification suggested in step A. We see a reduction in read-pulse duration. ($N$: neurons, $\alpha$: alpha-function gen., #: RRAM crossbar, $P$: peak-detector)

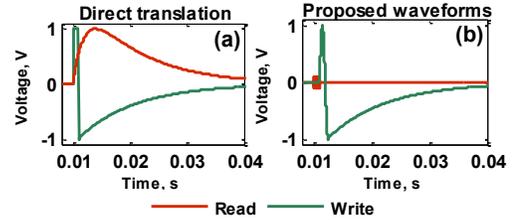

Figure 3: (a) Read and write-pulses applied on the RRAM array translated directly from algorithm; (b) read-write-pulses modified to enable simultaneous read and write for linear RRAMs.

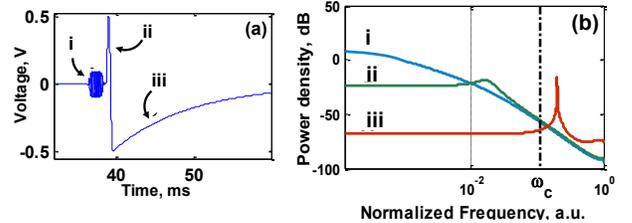

Figure 4: (a) Pre-synaptic terminal's signal and its components, applied immediately in response to a spike; (b) Power spectral density of the three components. The filter cut-off is around the band centered at $\omega_C$.

#### B. Using FDM for read vs. write-pulses

To enable FDM, we perform two steps. First, we replace the short read-pulse (described in the previous sub-section) with short high-frequency sinusoids. Second, we use a smoother form of the write-pulse, eliminating all sharp variations. Each time a neuron spikes, a short sinusoidal pulse is applied, with a short time-offset before the extended write-pulse Thus, the original extended read-pulse ($\alpha$-function) and write-pulse (to enable STDP) (Figure 3a) is converted into short, high-frequency read and long, low frequency write-pulse (Figure 3b). This allows the high-frequency read-pulse to be separated from lower frequency write-pulse with an appropriate filter, placed afterwards (Figure 4).

#### C. Selective Read-current extraction

After filtering, the read-pulse produces a sinusoidal current proportional to the instantaneous *small-signal* conductance. The zero-mean sinusoidal output is rectified and passed through a peak-detector (shown in gray in Figure 2) to extract the amplitude as pulse output, whose amplitude is dependent upon the synaptic conductance. This pulse drives the (LTI) synaptic spike response generator ($\alpha$-function), whose output then drives the LIF neuron to produce post-neuronal spikes.

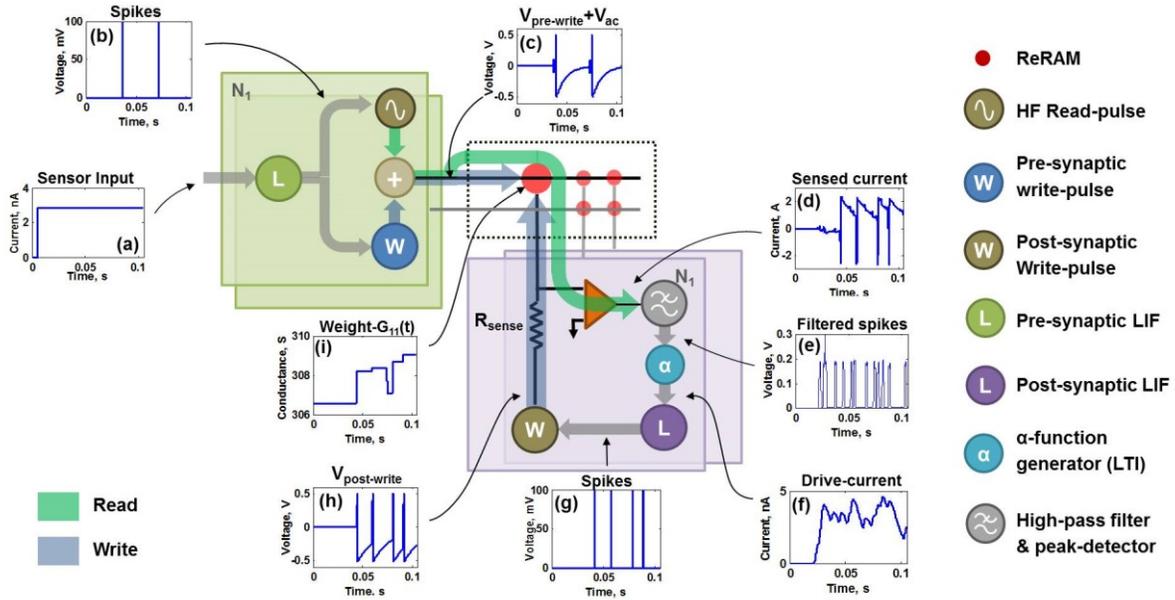

Figure 5: High-level schematic of the system and constituents, for a two-layers feed-forward SNN based on suggested scheme; (a) Input being provided from the sensor which the drives the LIF unit; (b) voltage is spiked each time membrane potential crosses the threshold; (c) a 'time since last spike' dependent waveform is generated each time spiking occurs. To this waveform, brief fast-varying sinusoid voltage is added; (d) Current through the current sense resistance; (e) High pass filter followed by an envelope detector transform sinusoids into rectangular voltage pulses; (f) These rectangular pulses drive the α-function generator, which then drives the second layer LIF unit; (g) The second layer neuron undergoes similar process as the first layer neurons, generating write-pulses shown in (h); (i) RRAM's weight changes whenever the voltage across it crosses its threshold.

The LTI approximation requires that RRAM being used as a synapse follows an "ideal" linear current-voltage relationship. We study the effect of non-linear DC I-V on an SNN performance in Section IV.D.

## IV. RESULTS

To demonstrate the proposed scheme, we base our SNN on a mathematical model discussed in [23]. We used *NG-SPICE-25* to describe and simulate the network. First, we discuss, in details, our approach to define the components. Then we validate the reading and writing processes by randomly exciting a 1×1 array and determining the error. Following that, the complete SNN, on SPICE, is tested against the mathematical model. Lastly, we study effects two non-idealities of an RRAM on the SNN developed using proposed scheme: dependence of write voltage thresolds on conductance and non-linear DC I-V.

### A. The network and its components

The SPICE network shown in Figure 5, consists of analog currents based on population encoded data (Fig. 5a) to serve as input current for LIF neuron (symbol 'L') to produce spikes (Fig. 5b). These spikes drives pre-synaptic high-frequency read (symbol 'R') and low frequency write (symbol 'W') combined by an adder (symbol '+'). This combined read-write-pulse from pre-neurons (Fig. 5c) is applied to the crossbar array with RRAM as a synapse. The output current is sensed across a small resistor and the resultant voltage is amplified (Fig. 5d). Then, the high-frequency read-pulse is filtered out and rectified to form spikes (Fig. 5e), which drives

$\alpha$-function generator (symbol '$\alpha$') to produce the input current for LIF neuron (symbol 'L') (Fig. 5f). The neuron generates spikes (Fig. 5g), which drives the post-synaptic write-pulse (symbol 'W') to generate write-pulses (Fig. 5h). This is fed back into the crossbar array such that superposition of pre- and post-neuron write-pulses produce STDP (Fig. 5i). We note that the pre- and post-neuronal write-pulses need to be filtered-out to enable read operation. Each of the circuit elements are described below.

#### 1) LIF Neuron circuit:

Several circuits have been proposed that model an LIF neuron of Eq. 1 [11], [17], [25], [26]. These circuits essentially contain a capacitor in parallel with a conducting element and a reset switch. The capacitance is discharged when capacitor's charge crosses a threshold.

#### 2) Memristor based synapse:

We developed two models for the RRAM, in SPICE: (1) an *ideal* RRAM model (similar to [27]) that mimicked the

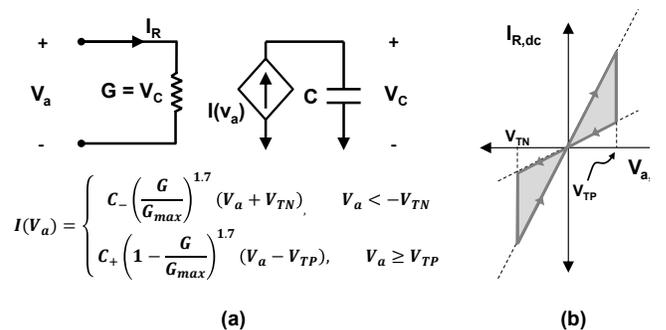

$$I(V_a) = \begin{cases} C_-\left(\dfrac{G}{G_{max}}\right)^{1.7}(V_a + V_{TN}), & V_a < -V_{TN} \\ C_+\left(1 - \dfrac{G}{G_{max}}\right)^{1.7}(V_a - V_{TP}), & V_a \geq V_{TP} \end{cases}$$

Figure 6: (a) *Ideal* RRAM's model that mimics the synapse in Eq. 4. Parameters are given in Table 2; (b) DC I-V schematic for the ideal RRAM model. Grey region represents range of conductance.

synapse in Eq. 4 and (2) a model of a *realistic* memristor i.e. HfO$_2$-RRAM that mimics the DC I-V of the HfO$_2$. In a memristor, the conductance ($G$) change only when applied bias $V_a$ exceeds a threshold in positive ($V_{TP}$) and negative ($V_{TN}$) polarity, while it otherwise remains unchanged. To model this behavior, the ideal RRAM model consists of a linear conductor ($G$), a capacitor ($C$) and a dependent current-source ($I$) shown in Figure 6a. $G$ depends linearly on voltage across the capacitor ($V_C$). A dependent current-source pumps charge into and out of $C$ only if (i) the applied voltage's magnitude ($|V_a|$) is above the write-thresholds' magnitude ($|V_{TP}|$ or $|V_{TN}|$) and (ii) the conductance is within a fixed range (0 and $G_{max}$) i.e. capacitance voltage is between two limits. If read using voltages below the thresholds, the DC I-V of the model would look like that of an ideal resistance, with a slope depending on $V_C$ (Figure 6b). The amount of charge pumped for a given $\Delta t$ (and hence the LTP or LTD rate) is controlled by the width and strength of the write-pulse (Figure 1f-h). HfO$_2$ RRAM's DC I-V model is essentially same, except that $V_{TP}$ and $V_{TN}$ depend on $V_C$, as described in Section IV.D. The SPICE model is shared on [28].

*3) Alpha-function generator*

Alpha-function generator is an LTI circuit whose impulse response is given in Eq. 3 and the circuit in shown in Figure 7. Impulse-response simulated in SPICE is shown in Figure 7b-c.

*4) Neuron write-pulse generator*

This unit generates write-pulses similar to the ones described in [7]. The pulse has a sharp positive spike followed by a slower negative timing part, latter being derived from the STDP rule being applied on the network. However, unlike the pulses in [7], it's smoothened to minimize filter's complexity. The following equation describes waveform of the write-pulse we used (parameters' values in Table 2):

$$w_\pm(t) = \begin{cases} 0.5 \sin\left(2\pi \dfrac{t}{T_w}\right), & 0 < t < \dfrac{3T_w}{2} \\ -0.5\left(\exp\left(-\dfrac{t - \dfrac{3T_w}{2}}{\tau_\pm}\right)\right), & t \geq \dfrac{3T_w}{2} \end{cases} \quad (5)$$

Here, +(-) represents the pre- (post-) synaptic write-pulse. The response is restarted each time a spike is applied to the generating unit i.e. it is not an LTI system. A simple circuit design for write-pulse generation is shown in Figure 8, where a voltage-controllable resistance (e.g. n-MOSFET) charges/discharges a capacitor a different timescales that depends upon the n-MOSFET resistance. A bipolar input pulse is applied to the S/D terminal, immediately after the driving neuron's spike, while a positive pulse is applied at the gate. For the positive S/D input, and positive input on gate, the n-MOSFET turns on strongly to charge the capacitor quickly to positive bias. When the S/D input sign changes to negative, the positively charged capacitor becomes the source. The n-MOSFET remains strongly turned on to negatively charge the capacitor. After the gate turns off, a slow discharge occurs through the highly resistive n-MOSFET. The input-output is validated in SPICE (Figure 8b).

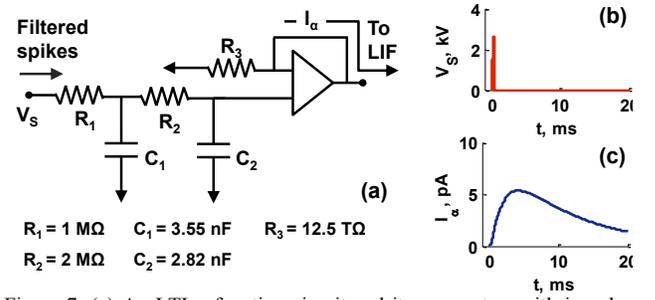

Figure 7: (a) An LTI α-function circuit and its parameters with impulse response Eq. 3; (b) Impulse-response of the α-function generator, simulated in SPICE. The output current strength (ii) is proportional to the input impulse in (i). The circuit was simulated with 180nm TSMC's MOSFET model.

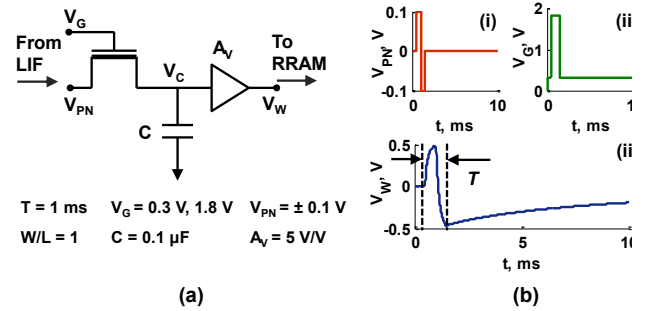

Figure 8: (a) Elements in the proposed write-pulse generation circuit. It approximates the output given in Eq. 5; (b) SPICE waveforms of (i) $V_{PN}$ and (ii) $V_G$ and (iii) output of the circuit. At each spiking instant, high $V_G$ and bi-polar $V_{PN}$ are applied that regenerate the write-pulse. The circuit was simulated with 180nm TSMC's MOSFET model.

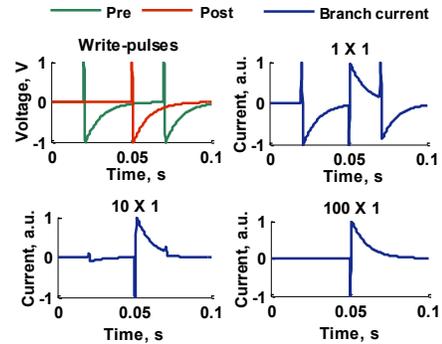

Figure 9: Increasing obscurity of pre-synaptic write-pulse current as crossbar's row size increases (same applies to read-current.) To enable extraction, reading is done in a higher frequency domain.

*5) High-pass filter*

We used a behavioral model of a high-pass filter to extract the read-pulse's high-frequency component from the low-frequency pre- and post-synaptic write-pulses. Within the pre-synaptic read- and write-pulse, the read-pulse is $k$-times smaller in amplitude than the write-pulse. For an $m \times n$ array, generally, $m$ (number of inputs) significantly exceeds $n$ (number of classes) i.e. $m \gg n$. The post-synaptic write-pulse produces a current that feeds $m$ parallel synapses of an average weight $G_o$ while the pre-neuron's read-pulse produces a current that passes through just one synapse to reach the sense-resistor. Thus, the post-synaptic write-pulse produces a current that is $m$ times as large as that of the pre-synaptic write-pulse (Figure 9). Thus, the low-frequency post-synaptic

TABLE II

PARAMETERS USED IN SPICE

| Component/Signal | Parameter | Value |
|---|---|---|
| Ideal RRAM | $C$ | 1 F |
| | $V_{TP}$, $V_{TN}$ | 0.5 V, -0.5 V |
| | $G_{max}$ | 700 S |
| | $B$ | $1.94 \times 10^4$ |
| | $p$ | 1.7 |
| Read-sinusoid | Frequency | 0.1ms |
| | Duration | 2.5ms |
| Write-pulse | $T_w$ | 2ms |
| | $\tau_+$, $\tau_-$ | $10ms$, $20ms$ |
| High-pass filter | $\omega_{3dB}$ | $2\pi \times 10^4$ rad/s |
| | Order | 8 |

write-current is $m.k$ times larger than the read-current. The filter must reject this low-frequency post-neuronal write component by attenuating by a factor of $r$ and pass the high-frequency pre-neuronal read-pulse, i.e. $m.k.r \ll 1$.

A filter has primarily two parameters: (i) cut-off frequency and (ii) order. Cut-off frequency is placed at the frequency of the sinusoid (Figure 4). The order of the filter is dependent on *ratios* of (1) strength and (2) frequencies of high and low frequency signals. Sinusoid's amplitude, which determines (1), has to be smaller than the RRAM's writing threshold. In our simulation's it was set to 0.1V (20% of the ideal RRAM's writing threshold). The ratio of frequencies can be made larger by using faster sinusoids. A numerical simulations challenge in SPICE transient analysis is that the time-step should be much smaller (~ 0.1×) of the fastest signal in the circuit, which, in our SNN, is the reading sinusoid. Thus, in order to limit the simulation time, we used a sinusoid that is only ten times as fast the fastest component of the write-pulse, as seen from Table 2. To model an "ideal" high-pass filter, we placed in cascade eight stages of RC elements, giving an order of eight. This limited the read-error to less than 2% (Figure 10a). In an actual hardware, higher read-pulses can reduce filter's order. However, our ideal demonstration is sufficient to demonstrate the concept.

### B. 1×1 array validation

To test reading and writing process occurring simultaneously, we developed in SPICE a 1×1 array, consisting of an input neuron connected to an output neuron through an RRAM based synapse. We used similar current levels, as used in the mathematical model in [23], to drive the neurons and validate the network.

For read-process, the input and output neurons were driven randomly and independently (akin to spontaneous spiking on neurons), allowing the weight of the synapse in-between to evolve freely in its range. We evaluated the reading process by comparing the filter output (ideally, proportional to the sampled conductance at the spiking instant) and the actual conductance. Histogram in Figure 10a shows the distribution of read-error, per-cent of the actual weight. The maximum error is around 2.0%, which implies that reading process is taking place in parallel, robustly and undisturbed by the write-process, for random spike times of the pre- and post-neurons.

To validate the write-process, we noted the change in conductance ($\Delta G$) for every pair of pre and post spikes ($\Delta t$) to produce an accurate STDP in the circuit compared to mathematical model (Figure 10b). A mean absolute error in $\Delta G$ of 2.4% was observed. Thus, this process too, is assumed free of disturbances due to the simultanoues array-reading process.

### C. Iris classifier network with an ideal synaptic model

We demonstrate a feed-forward SNN classifier network based on the Fisher Iris dataset, using the same procedure reported in details in [23]. In brief, each of the 4 attribute is first normalized, and then population coded into 4 Gaussian receptive fields to produce a 16×3 network, where the 3 output neurons correspond to 3 classes [23]. First, 30% of Fisher's Iris data is used for training the network by the STDP rule. Each sample is input to the network for 100ms. An epoch consists of sequential input of all training samples. Next, all the samples are used for classification based performance test.

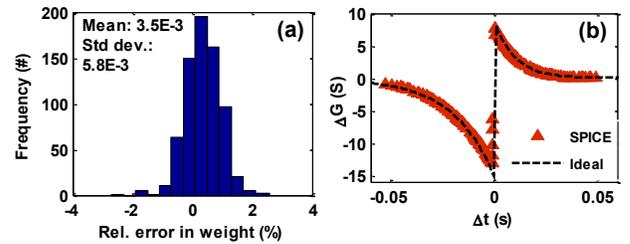

Figure 10: (a) Error distribution for the read-process done with filtration of sinusoidal current component through an ideal RRAM model; (b) Weight change for various inter pre-post spike intervals, for an ideal RRAM model. Saturation effects on weight change were disabled (setting $p = 0$ in Eq. 4) for validation. Mean absolute error is 2.4%

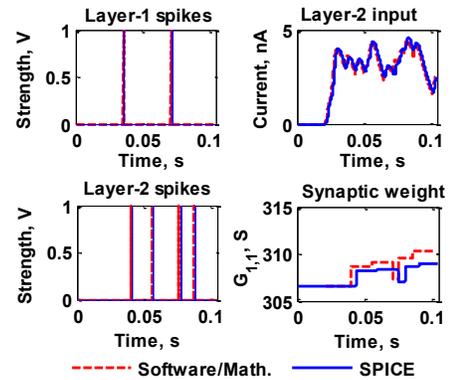

Figure 11: Essential signals associated with the first synapse of the 16×3 Iris classifier, in software (ideal) and SPICE based models. There is a slight mismatch in the weight evolution due to waveform *smoothening*.

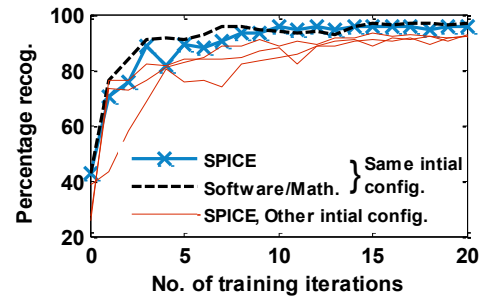

Figure 12: Classification accuracy as training progresses. Mathematical model's performance (dashed curve) is the benchmark

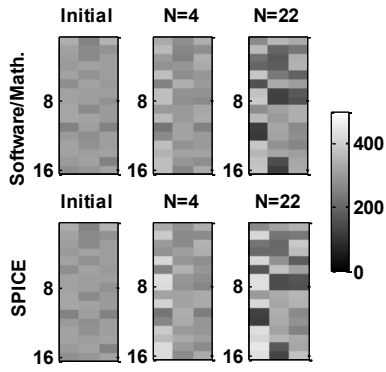

Figure 13: Grey-map for initial and final weights. As training progresses, weight approach similar values, thus validating the SPICE model of the circuit implementing the proposed scheme.

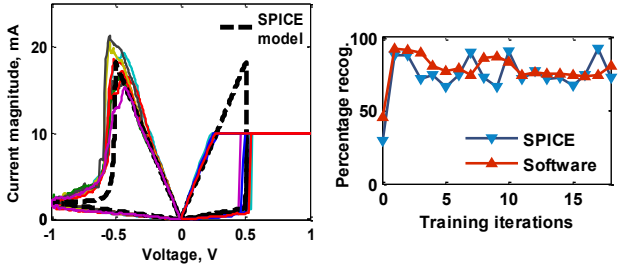

Figure 14: (a) Measured DC I-V of HfO$_2$ RRAM device and simulated DV I-V of device's behavioural model; (b) Percentage accuracies for SPICE and ideal Iris-classifying SNN using HfO$_2$ DC I-V model, as training progresses.

An *ideal* synapse model, described earlier, is employed.

The time evolution of input and output spikes and the $\alpha$-function current in SPICE are identical as the mathematical model (Figure 11). A lower weight change is observed due to the smoothening of the waveforms but this only lowers the learning rate but not the learning performance. The network's learning performance with ideal RRAM synapse is given in Figure 12. After training for twenty epochs, an accuracy of 97.33% is achieved in SPICE implementation, which is identical to software implementation.

*D. Iris classifier network with HfO$_2$ memristive synaptic model: Effect of Non-Ideal Resistive Switching*

Next, the *ideal* RRAM model was replaced by an area-scaled, realistic HfO$_2$ RRAM based compact SPICE model that mimics experimental DC I-Vs characteristics (Figure 14a). The device was modeled by modifying the $G$, $V_{TN}$ and $V_{TP}$ in Figure 6a as follows:

$$G(V_C) = \frac{31.5}{|V_C|} + \frac{0.127}{|V_C|(V_C + 0.48)^2} \quad (6a)$$

$$V_{TP} = 0.5 - 10^{-2}(V_C + 0.5) \quad (6b)$$

$$V_{TN} = V_C \quad (6c)$$

The dependence of $V_{TN}$ on $V_C$ or $G(V_C)$ degrades the maximum learning performance to 92% (Figure 14b). In fact, we observe that the classification accuracy does not stabilize as learning progresses, even though the learning rate is similar to the mathematical SNN. A similar degradation was observed in the mathematical (Figure 14b) and the two-array scheme based networks [22] that employed the same model. This implies that SNN performance degrades due to the conductance dependence of $V_{TP}$ or $V_{TN}$ and not because of the SPICE network.

*E. Effect on performance due to RRAM's I-V non-linearity: Effect of non-ideal conductance*

For a linear RRAM [14], [21], the small-signal conductance is same as the DC conductance of the device. However, some RRAMs have inherently a non-linear (quadratic) DC I-V relationship. For such devices [29], [30], the small-signal conductance of RRAM (and hence the filtered branch-current) depends on the read-bias/quiescent voltage. This is because a zero mean read-bias is offset by net write-pulses' voltage, being applied across the devices at the reading instant. Approximately, the current being fed-forward depends on the derivative of DC I-V curve (small-signal conductance). For quadratic RRAMs, small-signal conductance is linearly proportional to the bias-voltage.

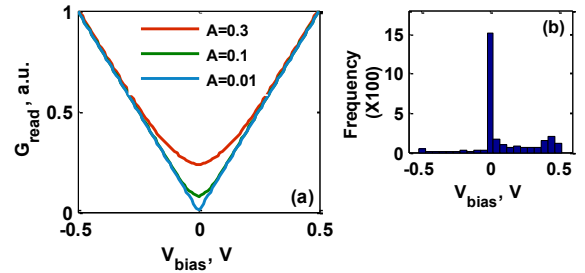

Figure 15: (a) Normalized small-signal conductance for quadratic RRAM model brought to a fixed iso-voltage DC conductance ($v_{bias}$). The y-intercept increases as read-amplitude is increased; (b) Distribution of Q-point/bias voltage for the first epoch. Similar distributions peaking at 0V are observed (not shown) for subsequent epochs.

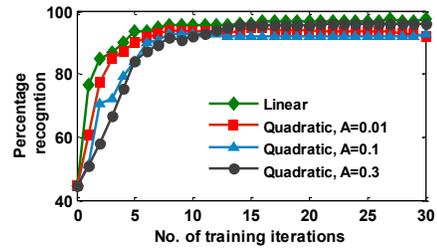

Figure 16: Percentage recognition increases as the number of iterations increases.

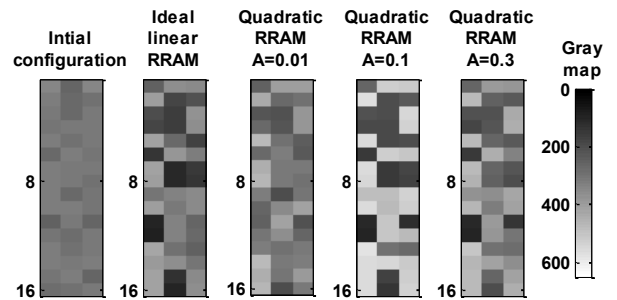

Figure 17: Initial and final (45 epochs) iso-voltage DC conductance-maps for linear and quadratic RRAM models. It is seen that if sinusoid's amplitude is big enough, final weights approach similar values regardless of the RRAM's I-V type.

TABLE III
PROPOSED SCHEME – TRAINING PERFORMANCE

| RRAM I-V type | Amplitude (V) | Classification accuracy (Max.) |
|---|---|---|
| Linear (ideal) | - | 97.3% |
| Linear (HfO$_2$, I-V matched) | - | 92% |
| Quadratic | 0.01 | 91.3% |
| | 0.1 | 93.0% |
| | 0.3 | 96.0% |

To explore the effect of bias dependence of the small-signal conductance, we replaced the *ideal* read-equation in the SNN training algorithm with the following:

$$I(t) = G_{ij}\, g_{\text{read}}(v_{bias}, A)\, \alpha_{\tau_1, \tau_2}(t) \quad (7)$$

where, $G_{\text{read}}(v_{bias}, A)$ is the effective (essentially average) conductance of an ideal RRAM calculated at offset bias/Q-point ($v_{bias}$) about an amplitude ($A$) of the read-pulse. The various $G_{\text{read}}(v_{bias})$ is plotted in Figure 15a. At $|v_{bias}| \gg A$, $G_{\text{read}}(v_{bias})$ is A independent. When $|v_{bias}| \ll A$, the $G_{\text{read}}(v_{bias})$ increases with $A$. Thus, the quadratic device appears more ideal (i.e., constant $G(v_{bias})$) over a larger $v_{bias}$ range (proportional to $A$). The offset's distribution during learning process, in Figure 15b, shows that a small fraction of read-pulses had a voltage offset (akin to $v_{bias}$ in Fig. 15a). Thus, high $A$ should enable more "ideal" behavior. Figure 16 plots the classification accuracy of the network for various amplitudes as learning progresses. As expected, learning performance approaches the ideal one's as the sinusoid's amplitude is increased (Table 3). The amplitude was not increased beyond 0.3V, since the RRAM model has a threshold of 0.5V and an unconstrained increase might disturb the writing process. Figure 17 plots the initial and final states for an ideal linear and quadratic RRAM. We see that final weight configuration is similar for all cases suggesting that the weight evolution is similar despite the different non-linear read processes. Thus, non-ideal conductance of RRAM produces software-equivalent results in the circuit implemental despite some LTI based circuits being used in the system.

## V.  BENCHMARKING

Several researchers have reported their implementation of biologically inspired neural network using both SRAM and RRAM arrays [11], [14], [17]–[19], [21], [22], [31] which are compared in Table 4, from the perspective of enabling bio-mimetic neural networks – essentially asynchronous spiking for real-time learning and recognition. Some of these works are *discrete*-time systems and spiking is synchronized with a *common clock* with reading and writing separated in time [11], [17], [18]. Although several groups also focus on *arbitrarily timed or asynchronous* spiking networks [14], [19], [21], [31] the dilemma is solved by switching off the reading process, during application of the write-pulse. However, (1) this requires spike-controlled switches between layers of neurons and (2) if the total number of spikes from the pre-synaptic layer is large, current integration at the post-synaptic neural layer may get disturbed due to frequent disablement of reading. In [22], we split reading and writing processes physically by using two RRAM arrays – one array is used for reading and the other for writing. As learning progresses, write array is written onto. Periodically, write array's content is transferred onto the read array. This requires an additional circuitry to transfer weight and thus is not really asynchronous. In comparison, the presently proposed scheme yields a clock-less and arbitrarily timed SNN with on-chip learning and does not involve disabling of reading while writing occurs.

TABLE IV
SCHEMES FOR STDP BASED REAL-TIME LEARNING ON SNN HARDWARE WITH MEMRISTIVE SYNAPSES

| | Key features | | | | |
|---|---|---|---|---|---|
| | Change in weight presented | Proposed synapse | Simultaneous read-write | Clock-less read-write differentiation | Strategy to negotiate read-write Dilemma |
| *Seo et al.* [11] | Digital, multi-level | 8T-SRAM | No | No | Multiple clock phases required for independent learning and recognition phases |
| *Kim et al.* [17] | Analog, using PWM | 1R memristor | No | No | |
| *Cruz-Albrecht et al.* [18] | Multi-state, using timer circuit | 1R memristor | No | No | |
| *Serrano-Gotarredona et al.* [30] | Analog, using write-pulse superposition | 1R memristor | Yes | Yes | Additional inter-layer communication needed |
| *Wu et al.* [19] | Analog, using write-pulse superposition | 1R memristor | No | Yes | Reading disabled during write-pulse application |
| *Wang et al.* [14] | Binary, using write-pulse superposition | 2T-1R memristor | No | Yes | Reading disabled during write-pulse application |
| *Pedretti et al.* [21] | Binary, using write-pulse superposition | 1T-1R memristor | No | Yes | Reading disabled during write-pulse application |
| *Shukla et al.* [22] | Analog, using write-pulse superposition | 1R memristor | Yes | Yes | Double area needed for separating reading and writing |
| *This work* | Analog, using write-pulse superposition | 1R memristor | Yes | Yes | FDM used for interlayer communication |

## VI. CONCLUSION

For hardware SNNs employing crossbar RRAM arrays as synaptic-array, the simultaneous and asynchronous read-write dilemma is a key challenge. We propose performing read and writing processes in different frequency domains to minimize interference between read and write processes. To demonstrate our idea, we translate a mathematical SNN based Iris-classifier into SPICE. Three key results are shown. First, an ideal RRAM based synapse enables mathematically equivalent training performance (i.e., 97.3% accuracy) in the circuit implementation. Second, we explored the effect of *non-ideal resistive switching* which involved dependence of write-voltage threshold on the conductance-state. We simulated our SPICE network using SPICE model of $HfO_2$ RRAM by matching experimental I-Vs. After training, the learning performance was found to degrade to a maximum of 92% accuracy for both mathematical and SPICE networks. This implies that the RRAM ideality is critical for performance. Finally, we show the effect of non-ideal conductance in RRAMs. RRAMs with non-linear I-V characteristics perform similarly as a linear RRAM, which underlines the robustness of the system despite having component that use the LTI approximation. Thus, a clock-less scheme for SNN is demonstrated using FDM that enables software-equivalent on-chip learning. Such equivalence enables a direct translation of software algorithms to hardware.

## REFERENCES

[1] J. von Neumann, *The Computer and the Brain*. Yale University Press, 1958.
[2] W. Maass, "Noisy spiking neurons with temporal coding have more computational power than sigmoidal neurons," *Adv. Neural Inf. Process. Syst.*, vol. 9, pp. 211–217, 1997.
[3] W. Maass and C. M. Bishop, *Pulsed Neural Networks*, vol. 275. MIT Press, 1999.
[4] S. B. Furber *et al.*, "Overview of the SpiNNaker system architecture," *IEEE Trans. Comput.*, vol. 62, no. 12, pp. 2454–2467, 2013.
[5] C. Johansson and A. Lansner, "Towards cortex sized artificial neural systems," *Neural Networks*, vol. 20, no. 1, pp. 48–61, Jan. 2007.
[6] H. Esmaeilzadeh, E. Blem, R. St. Amant, K. Sankaralingam, and D. Burger, "Dark Silicon and the End of Multicore Scaling," in *38th International Symposium on Computer Architecture*, 2011, vol. 39, no. 3, pp. 365–376.
[7] B. Rajendran *et al.*, "Specifications of Nanoscale Devices and Circuits for Neuromorphic Computational Systems," *IEEE Trans. Electron Devices*, vol. 60, no. 1, pp. 246–253, Jan. 2013.
[8] P. A. Merolla *et al.*, "A million spiking-neuron integrated circuit with a scalable communication network and interface," *Science (80-. ).*, vol. 345, no. 6197, pp. 668–673, Aug. 2014.
[9] B. V. Benjamin *et al.*, "Neurogrid: A Mixed-Analog-Digital Multichip System for Large-Scale Neural Simulations," *Proc. IEEE*, vol. 102, no. 5, pp. 699–716, May 2014.
[10] J. Schemmel, D. Brüderle, A. Grübl, M. Hock, K. Meier, and S. Millner, "A wafer-scale neuromorphic hardware system for large-scale neural modeling," in *Proceedings of 2010 IEEE International Symposium on Circuits and Systems*, 2010, pp. 1947–1950.
[11] J.-S. Seo *et al.*, "A 45nm CMOS neuromorphic chip with a scalable architecture for learning in networks of spiking neurons," in *2011 IEEE Custom Integrated Circuits Conference (CICC)*, 2011, pp. 1–4.
[12] G. Q. Bi and M. M. Poo, "Synaptic modifications in cultured hippocampal neurons: dependence on spike timing, synaptic strength, and postsynaptic cell type.," *J. Neurosci.*, vol. 18, no. 24, pp. 10464–10472, 1998.
[13] N. Panwar, B. Rajendran, and U. Ganguly, "Arbitrary Spike Time Dependent Plasticity (STDP) in Memristor by Analog Waveform Engineering," *IEEE Electron Device Lett.*, vol. 3106, no. c, pp. 1–1, 2017.
[14] Z. Wang, S. Ambrogio, S. Balatti, and D. Ielmini, "A 2-transistor/1-resistor artificial synapse capable of communication and stochastic learning in neuromorphic systems," *Front. Neurosci.*, vol. 9, no. JAN, pp. 1–11, 2015.
[15] D. Garbin *et al.*, "$HfO_2$-Based OxRAM Devices as Synapses for Convolutional Neural Networks," *IEEE Trans. Electron Devices*, vol. 62, no. 8, pp. 2494–2501, Aug. 2015.
[16] D. Purves and S. M. (Stephen M. Williams, *Neuroscience*, 2nd editio. Sunderland (MA): Sinauer Associates, 2001.
[17] Y. Kim, Y. Zhang, and P. Li, "A Reconfigurable Digital Neuromorphic Processor with Memristive Synaptic Crossbar for Cognitive Computing," *ACM J. Emerg. Technol. Comput. Syst.*, vol. 11, no. 4, pp. 1–25, 2015.
[18] J. M. Cruz-Albrecht, T. Derosier, and N. Srinivasa, "A scalable neural chip with synaptic electronics using CMOS integrated memristors," *Nanotechnology*, vol. 24, no. 38, p. 384011, 2013.
[19] X. Wu, V. Saxena, and K. Zhu, "Homogeneous spiking neuromorphic system for real-world pattern recognition," *IEEE J. Emerg. Sel. Top. Circuits Syst.*, vol. 5, no. 2, pp. 254–266, 2015.
[20] S. Ambrogio *et al.*, "Neuromorphic Learning and Recognition With One-Transistor-One-Resistor Synapses and Bistable Metal Oxide RRAM," vol. 63, no. 4, pp. 1508–1515, 2016.
[21] G. Pedretti *et al.*, "Memristive neural network for on-line learning and tracking with brain-inspired spike timing dependent plasticity," no. May, pp. 1–10, 2017.
[22] A. Shukla, V. Kumar, and U. Ganguly, "A software-equivalent SNN hardware using RRAM-array for asynchronous real-time learning," *2017 Int. Jt. Conf. Neural Networks*, pp. 4657–4664, May 2017.
[23] A. Biswas, S. Prasad, U. Ganguly, S. Lashkare, and U.

Ganguly, "A simple and efficient SNN and its performance & robustness evaluation method to enable hardware implementation," Dec. 2016, *Arxiv*: *https://arxiv.org/abs/1612.02233*.

[24] A. V. Oppenheim, A. S. Willsky, and S. H. Nawab, *Signals and Systems*. Prentice Hall, 1997.

[25] J. M. Cruz-Albrecht, M. W. Yung, and N. Srinivasa, "Energy-efficient neuron, synapse and STDP integrated circuits," *IEEE Trans. Biomed. Circuits Syst.*, vol. 6, no. 3, pp. 246–256, 2012.

[26] G. Indiveri, E. Chicca, and R. Douglas, "A VLSI array of low-power spiking neurons and bistable synapses with spike-timing dependent plasticity," *IEEE Trans. Neural Networks*, vol. 17, no. 1, pp. 211–221, 2006.

[27] Y. V. Pershin and M. Di Ventra, "SPICE model of memristive devices with threshold," vol. 22, no. 2, pp. 485–489, Apr. 2012.

[28] A. Shukla, "HfO2 RRAM - SPICE DC I-V model." [Online]. Available: *https://gist.github.com/adi2293/bfc0bdfc86950eef35773556d937f5ae*.

[29] P. Kumbhare *et al.*, "A Comprehensive Study of Effect of Composition on Resistive Switching of $Hf_xAl_{1-x}O_y$ based RRAM devices by Combinatorial Sputtering," *MRS Proc.*, vol. 1729, pp. 65–70, 2015.

[30] S. Park *et al.*, "Nanoscale RRAM-based synaptic electronics: toward a neuromorphic computing device.," *Nanotechnology*, vol. 24, no. 38, p. 384009, 2013.

[31] T. Serrano-Gotarredona, T. Masquelier, T. Prodromakis, G. Indiveri, and B. Linares-Barranco, "STDP and STDP variations with memristors for spiking neuromorphic learning systems," *Front. Neurosci.*, vol. 7, no. 7 FEB, pp. 1–15, 2013.